\PassOptionsToPackage{table}{xcolor}
\documentclass[manuscript,nonacm]{acmart}
\usepackage{multirow}
\usepackage[most]{tcolorbox}
\usepackage{enumitem}
\usepackage[compact]{titlesec} 
\usepackage{adjustbox}

\renewcommand\footnotetextcopyrightpermission[1]{}
\AtBeginDocument{%
  }

\makeatletter
\renewcommand\footnotetextcopyrightpermission[1]{\footnotetext{This manuscript has been authored by UT-Battelle, LLC, under contract DE-AC05-00OR22725 with the US Department of Energy (DOE). The US government retains and the publisher, by accepting the article for publication, acknowledges that the US government retains a nonexclusive, paid-up, irrevocable, worldwide license to publish or reproduce the published form of this manuscript, or allow others to do so, for US government purposes. DOE will provide public access to these results of federally sponsored research in accordance with the DOE Public Access Plan (https://www.energy.gov/doe-public-access-plan).}}
\makeatother

\newtcolorbox[auto counter]{finding}[1][]{%
    colback=blue!5,           
    colframe=blue!40,         
    boxrule=0pt,              
    leftrule=2mm,             
    sharp corners,            
    before skip=2pt,
    after skip=0pt,
    fontupper=\normalfont,    
}

\begin{document}

\title{Data Readiness for Scientific AI at Scale}

\author{Wesley Brewer}
\orcid{0000-0002-3639-3956}
\affiliation{%
  \institution{Oak Ridge National Laboratory}
  \city{Oak Ridge}
  \state{Tennessee}
  \country{USA}}
\email{brewerwh@ornl.gov}

\author{Patrick Widener}
\affiliation{%
  \institution{Oak Ridge National Laboratory}
  \city{Oak Ridge}
  \state{Tennessee}
  \country{USA}}
\email{widenerpm@ornl.gov}

\author{Valentine Anantharaj}
\affiliation{%
  \institution{Oak Ridge National Laboratory}
  \city{Oak Ridge}
  \state{Tennessee}
  \country{USA}}
\email{anantharajvg@ornl.gov}

\author{Feiyi Wang}
\affiliation{%
  \institution{Oak Ridge National Laboratory}
  \city{Oak Ridge}
  \state{Tennessee}
  \country{USA}}
\email{fwang2@ornl.gov}

\author{Tom Beck}
\affiliation{%
  \institution{Oak Ridge National Laboratory}
  \city{Oak Ridge}
  \state{Tennessee}
  \country{USA}}
\email{becktl@ornl.gov}

\author{Arjun Shankar}
\affiliation{%
  \institution{Oak Ridge National Laboratory}
  \city{Oak Ridge}
  \state{Tennessee}
  \country{USA}}
\email{shankarm@ornl.gov}

\author{Sarp Oral}
\affiliation{%
  \institution{Oak Ridge National Laboratory}
  \city{Oak Ridge}
  \state{Tennessee}
  \country{USA}}
\email{oralhs@ornl.gov}




\renewcommand{\shortauthors}{Brewer et al.}

\begin{abstract}
This paper examines how Data Readiness for AI (DRAI) principles apply to leadership-scale scientific datasets used to train foundation models. We analyze archetypal workflows across four representative domains—climate, nuclear fusion, bio/health, and materials—to identify common preprocessing patterns and domain-specific constraints. We introduce a two-dimensional readiness framework composed of Data Readiness Levels (raw to AI-ready) and Data Processing Stages (ingest to shard), both tailored to high performance computing (HPC) environments. This framework outlines key challenges in transforming scientific data for scalable AI training, emphasizing transformer-based generative models. 
Together, these dimensions form a conceptual maturity matrix that characterizes scientific data readiness and guides infrastructure development toward standardized, cross-domain support for scalable and reproducible AI for science.
\end{abstract}




\maketitle


\section{Introduction}

The performance of AI systems is fundamentally limited by the quality and readiness of the data they consume. While progress has been made in training scientific foundation models, transforming domain-specific datasets into AI-ready formats remains labor-intensive and fragmented. AI-ready data—cleaned, labeled, normalized, feature-engineered, and formatted for scalable training—is especially difficult to produce in scientific domains, some of which lack uniform preprocessing pipelines, metadata standards, and performance-aware formats.

We investigate the specific challenges of AI-readiness at leadership computing scale, focusing on four strategic domains—climate, fusion, bio/health, and materials science—where foundation models are emerging as key enablers of discovery~\cite{bodnar2025foundation, pyzer2025foundation}. In this context, we focus on preprocessing requirements for foundation models, with an emphasis on transformer-based generative models that are increasingly applied across scientific domains. Drawing on domain workflows and expert interviews, we identify common preprocessing patterns and propose a five-level readiness framework to classify datasets based on their proximity to being “ready-to-train.”
Our goal is to provide a taxonomy and practical insights to guide the development of reusable, scalable, and secure data readiness pipelines. While our analysis is informed by activities at the Oak Ridge Leadership Computing Facility (OLCF)~\cite{abraham20242023}, the patterns and recommendations are broadly applicable to other DOE facilities and initiatives like the NAIRR~\cite{nairr2023final}.

Scientific AI poses distinct challenges compared to commercial AI. Datasets are often sparse, high-dimensional, and expensive to generate~\cite{brewer2024scalable}, requiring high-precision formats~\cite{kashi2024mixed} and adherence to physical constraints~\cite{cranmer2020lagrangian, meng2025physics}. These demands necessitate facility-aware AI-readiness pipelines.
Scale adds further complexity. The ClimaX model, for example, was trained on over ten terabytes of climate data~\cite{nguyen2023climax}. Efficient training at this scale requires high-throughput, parallel file I/O. These scalability challenges reinforce the need for structured frameworks that can guide infrastructure design and workflow development for scientific AI.



Our contributions include: (1) a cross-domain investigation of data readiness practices, (2) a five-level AI-readiness classification with examples, (3) identification of key infrastructure and governance challenges, (4) actionable recommendations, and (5) a two-dimensional readiness framework.

This paper presents an initial step toward formalizing AI-readiness workflows at leadership-scale computing facilities. Our goal is to provide practical guidelines and a structured readiness framework to help researchers and facility operators transform scientific datasets into formats optimized for scalable, reproducible AI model training. Further iteration and validation are needed to refine and operationalize these insights across domains.

\section{Background}

To motivate our readiness framework, we outline: (1) typical preprocessing patterns, (2) defining characteristics of scientific datasets, and (3) related community efforts that inform cross-domain readiness.

\begin{figure*}
    \centering
    \includegraphics[width=0.9\linewidth]{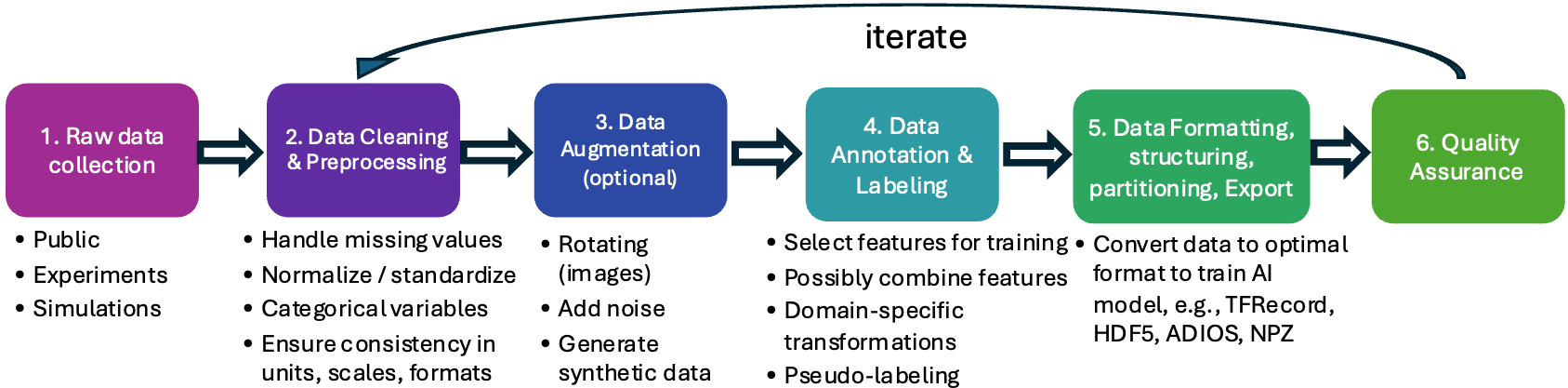}
    \caption{General steps to transform data from raw format to “AI-ready”. This process may vary among domains.}
    \label{fig:process}
\end{figure*}

\subsection{Common Preprocessing Patterns}

A key challenge in data readiness for AI is the transformation of raw scientific data into formats suitable for efficient, large-scale model training. While the specific steps may vary across domains, most workflows involve a common sequence of data cleaning, normalization, labeling, feature engineering, and format conversion. These steps must also be adapted for high-throughput, parallel data ingestion in HPC environments.

Figure~\ref{fig:process} summarizes the typical stages involved in this transformation. It illustrates how data originating from simulations, experiments, or public repositories is incrementally processed to achieve an AI-ready state—defined by sharded storage in binary formats such as HDF5, ADIOS~\cite{lofstead2008flexible}, or TFRecords. 
Scientific data is sourced from either public websites, experiments, or simulations. 
Common preprocessing tasks include handling missing values, \textit{normalizing} by mean and standard deviation, managing categorical variables, and ensuring consistent units and formats~\cite{ElMorr2022}.
In cases, typically concerning images, where scientific datasets contain an insufficient number of samples, certain \textit{data augmentation} techniques may be employed to increase the number of samples (such as rotating images, adding noise, and generating synthetic samples). 
Some datasets require manual or automated labeling, and when only a portion of the data is labeled, \textit{semi-supervised learning} methods can leverage both labeled and unlabeled samples. A common strategy in this setting is \textit{pseudo-labeling}, where model predictions on unlabeled data are iteratively treated as labels to improve training~\cite{kage2024review}. 
The next step is generally performing some level of \textit{feature engineering} to select the most informative set of features or combination of features on which to train. Then, the data should be split into train, test, and validation sets, and finally exported in a standard compressed and sharded format. This pipeline is inherently iterative: data preparation outcomes inform subsequent model training, and model performance provides feedback that triggers further data refinement and augmentation. 

\subsection{Scientific Data Characteristics}
\label{sec:scientific_facility}

Scientific AI diverges from commercial AI not only in its data sources but in the rigor of its requirements. Many scientific datasets stem from high-fidelity simulations or costly experiments and are often accompanied by limited or inconsistent labels. In contrast to widely available, user-generated datasets, scientific data requires significant preprocessing to achieve uniformity, compliance, and reproducibility.

Accuracy is critical—engineering and physics-based models often demand 32-bit or 64-bit floating-point precision to preserve numerical stability and physical realism~\cite{kashi2024mixed}. 
Applications such as self-driving systems, virtual sensors in aerospace, or surrogate models in computational physics must adhere to domain-specific constraints such as conservation laws and boundary conditions~\cite{raissi2019physics}.

These challenges place significant demands on AI infrastructure at leadership facilities. Data must be prepared at petascale or exascale volumes, support secure and auditable workflows, and interface efficiently with GPU-accelerated AI training pipelines. As scientific AI continues to evolve, readiness frameworks must therefore go beyond generic AI data pipelines to accommodate the unique technical and operational requirements of large-scale scientific computing.

\subsection{Related Work}

Hiniduma et al.~\cite{hiniduma2025data} provide a comprehensive 360-degree survey of data readiness across multiple dimensions, including data quality, ethical compliance, infrastructure, and governance. Their framework establishes a broad conceptual foundation for understanding readiness in AI workflows. Building on this, we focus specifically on the operational challenges of transforming scientific datasets into AI-ready formats at leadership scale.

Multiple community efforts have also emphasized the need for structured, high-quality datasets in domain-specific contexts. For example, the AI4ESP initiative~\cite{doe2022ai4esp} identified AI-readiness as a top priority for Earth system modeling, calling for standardized and quality-checked datasets. Similarly, the Fusion Machine Learning (ML) workshop \cite{fusion2019report} highlighted that fusion scientists spend the majority of their time curating data due to inconsistent formats and missing labels.


Our work builds on these efforts by offering a cross-domain classification of AI-readiness levels and patterns. Unlike prior domain-specific studies, we present a unified framework that spans climate, fusion, bio/health, and materials science, informed by leadership-scale workflows. This broad perspective enables the identification of shared bottlenecks and readiness strategies that can be generalized to future scientific foundation model initiatives.

\section{Readiness Patterns Across Domains}


We surveyed four strategic scientific domains—climate, fusion, bio/health, and materials—to understand the current status and challenges of AI-readiness. These domains were selected because they represent strategic priorities within DOE leadership computing facilities, cover diverse data modalities and preprocessing requirements, and uniquely demonstrate AI-readiness challenges and opportunities. Our analysis draws on interviews with domain scientists at ORNL, as well as representative projects and datasets within each field. We emphasize common preprocessing bottlenecks and indicators of readiness, and note domain-specific constraints such as privacy or provenance where applicable. We summarize representative datasets, along with their preprocessing pipelines and domain-specific challenges, in Table~\ref{tab:challenges}.

To distill common patterns in AI-readiness, we selected one representative workflow from each of the four domains. These ``archetypes'' capture recurring preprocessing and infrastructure needs, which we later formalize into a five-level readiness framework.


\subsection{Climate}

Climate science data collections span simulations (e.g., CMIP), satellite observations (e.g., MODIS), ground-based measurements (weather radar) and hybrid reanalysis products (ERA5). Preprocessing challenges include spatial/temporal alignment, normalization, and heterogeneity across data sources. Foundation models like ORBIT \cite{wang2024orbit} and ClimaX \cite{nguyen2023climax} depend on converting community standard formats, such as (encoded) Gridded Binary (GRIB) and (self-describing) Network Common Data Form (NetCDF), to sharded NumPy files with standardized features. The AI4ESP report~\cite{doe2022ai4esp} explicitly prioritizes creating ``AI-ready datasets'' with benchmarks and quality-controlled preprocessing pipelines.

\small
\begin{table*}[h]
\centering
\caption{Representative datasets, preprocessing pipelines, AI methods, modalities, and readiness challenges across domains}
\label{tab:challenges}
\adjustbox{max width=\columnwidth}{
\begin{tabular}{|p{1.5cm}|p{2.5cm}|p{3cm}|p{2.5cm}|p{2.5cm}|p{3cm}|}
\hline
\textbf{Domain} & \textbf{Dataset/Source} & \textbf{Workflow Steps} & \textbf{Architecture} & \textbf{Modality} & \textbf{Readiness Challenges} \\
\hline

Climate &
\begin{tabular}[t]{@{}l@{}}
CMIP6 (ORBIT~\cite{wang2024orbit}) \\ satellite imagery \\ observations \\ ERA5 reanalyses 
\end{tabular} &
\begin{tabular}[t]{@{}l@{}}
• Normalize variables \\ • Resample grids \\ • Standardize outputs \\ • Shard to binary formats
\end{tabular} &
CNN, Transformer &
Spatial, Temporal grids &
\begin{tabular}[t]{@{}l@{}}
• Redundant fields \\ • Spatial misalignment \\ • Pipeline throughput
\end{tabular} \\
\hline

Fusion &
\begin{tabular}[t]{@{}l@{}}
IPS-Fastran~\cite{cianciosa2022adaptive} \\ FREDA~\cite{collins2023integrated} \\ DIII-D ML~\cite{degrave2022magnetic} \\ IMAS~\cite{romanelli2020code}
\end{tabular} &
\begin{tabular}[t]{@{}l@{}}
• Extract/align diagnostics \\ • Physics-based features \\ • Normalize shots \\ • TFRecord/HDF5
\end{tabular} &
Transformer, CNN, LSTM &
Time-series, Multi-channel signals &
\begin{tabular}[t]{@{}l@{}}
• Sparse/noisy data \\ • Limited labels \\ • Access restrictions
\end{tabular} \\
\hline

Bio/Health &
\begin{tabular}[t]{@{}l@{}}
TwoFold~\cite{hsu2023twofold} \\
C-HER~\cite{klasky2024va-pipeline} \\ 
Enformer~\cite{avsec2021enformer} \\ 
AlphaFold 2~\cite{jumper2021alphafold}
\end{tabular} &
\begin{tabular}[t]{@{}l@{}}
• One-hot encoding \\ • Anonymization \\ • Cross-modal fusion \\ • Secure sharding
\end{tabular} &
Transformer, CNN, GNN &
Sequences, Images, Tabular &
\begin{tabular}[t]{@{}l@{}}
• PHI/PII compliance \\ • Limited labels \\ • Format inconsistencies
\end{tabular} \\
\hline

Materials &
\begin{tabular}[t]{@{}l@{}}
OMat24~\cite{barrosoluque2024openmaterials2024omat24} \\ AFLOW~\cite{curtarolo2012aflow}
\end{tabular} &
\begin{tabular}[t]{@{}l@{}}
• Parse simulations \\ • Normalize descriptors \\ • Graph encoding \\ • Shard (ADIOS/JSON)
\end{tabular} &
Graph Neural Network (GNN)~\cite{pasini2024scalable} &
Graph structures &
\begin{tabular}[t]{@{}l@{}}
• Class imbalance \\ • Fidelity mismatch \\ • Graph complexity
\end{tabular} \\
\hline
\end{tabular}
}
\end{table*}
\normalsize

Machine learning pipelines for weather and climate deep learning models frequently process large, multivariate datasets derived from reanalyses datasets. A common preprocessing pattern across these deep learning models is the sequence:
\textit{download → regrid → normalize → shard}.

For instance, ClimaX preprocesses CMIP6 NetCDF files by interpolating spatial grids, normalizing each variable with computed mean and standard deviation, and storing the processed data as sharded NumPy (.npz) files. Similarly, Pangu-Weather regrids reanalysis data to uniform spatial resolutions, slices it into spatiotemporal patches, and shards it for efficient training. These patterns emphasize the need for standard grid alignment, resolution consistency, and scalable I/O.



\subsection{Fusion}

Fusion research leverages both simulation and experimental data to model plasma behavior, among other things. Data sources include IPS-Fastran \cite{cianciosa2022adaptive}, surrogate modeling efforts \cite{ghai2024surrogate}, and experimental facilities like DIII-D tokamak reactor~\cite{diii-d2024}. Fusion workflows are often fragmented, and access to experimental data can be restricted. The 2019 DOE fusion-ML workshop noted that “scientists spend upwards of 70\% of their time on data curation”~\cite{fusion2019report}, due to inconsistent formats, unlabeled data, and limited provenance metadata.

Fusion workflows are characterized by a combination of simulation and experimental data, requiring complex preprocessing to align, clean, and transform heterogeneous time-series signals. Across pipelines like DIII-D disruption prediction~\cite{degrave2022magnetic}, PyFusion edge turbulence analysis, and the Integrated Thermonuclear Experimental Reactor's (ITER) Integrated Modeling and Analysis Suite (IMAS)-based assimilation workflows~\cite{romanelli2020code}, we observe a consistent pattern:
\textit{extract → align → normalize → shard}.

For example, the DIII-D ML pipeline begins with shot-level data extraction via MDSplus data acquisition framework~\cite{fredian2018mdsplus}, slices high-rate sensor streams into fixed time windows, computes derivative-based features from diagnostics (e.g., coil voltages, plasma currents), and aggregates across shots before sharding into TFRecords~\cite{tfrecord}. Similarly, PyFusion regrids irregular plasma diagnostics and constructs 2D/3D feature maps standardized across campaigns.

This pattern requires domain-specific transformations (e.g., physics-informed features, time-alignment across diagnostics), and in some cases, regridding or interpolation across incompatible meshes (as in IMAS and XGC1). While simulation data may be well-structured, experimental data is often sparse, noisy, or poorly labeled.



\subsection{Bio/Health}

AI workflows in bio/health span molecular modeling, clinical imaging, and genomic sequence prediction. These datasets often include protected health information (PHI) and personally identifiable information (PII), requiring strict compliance with standards such as the Health Insurance Portability and Accountability Act (HIPAA) and policy frameworks and standards developed by the Global Alliance for Genomics and Health (GA4GH), and motivating the use of secure enclaves and federated learning strategies. Projects like AlphaFold 2~\cite{jumper2021alphafold}, Enformer~\cite{avsec2021enformer}, and ORNL’s TwoFold~\cite{hsu2023twofold} 
illustrate common preprocessing patterns, including sequence encoding, multimodal alignment, and privacy-preserving transformations. National reports have emphasized the need for stronger annotation standards, reproducibility infrastructure, and domain-specific data maturity models~\cite{gao2022ai_healthcare, zhao2022artificial}.

For instance, Enformer transforms DNA sequences via one-hot encoding, segments them into fixed-length tiles, and computes position-wise normalization statistics. AlphaFold 2 employs a complex preprocessing pipeline involving multiple sequence alignment (MSA), structural feature extraction, and intermediate caching for scalable model training~\cite{jumper2021alphafold}. In ORNL’s centralized health and environment repository (C-HER) effort~\cite{klasky2024va-pipeline}, multimodal clinical data—combining imaging, tabular, and vector data—requires both anonymization and integration across formats.



\subsection{Materials Science}

Materials datasets frequently stem from density functional theory (DFT) calculations and are well-suited to graph neural networks. Datasets like OMat24 \cite{barrosoluque2024openmaterials2024omat24} emphasize structural diversity and open access. At ORNL, HydraGNN \cite{pasini2024scalable} supports scalable foundation model training via ADIOS-sharded graph data. Challenges include normalization of atomic coordinates, data imbalance across materials classes, and integration of multi-fidelity simulation and experimental data.

Materials science pipelines increasingly rely on graph-based models to represent atomic structures, bonding interactions, and electronic properties. Common preprocessing steps involve parsing simulation outputs, normalizing atomic features, and constructing graph representations for training graph neural networks (GNNs). Projects such as HydraGNN~\cite{pasini2024scalable}, Open Materials 2024 (OMat24)~\cite{barrosoluque2024openmaterials2024omat24}, and AFLOW~\cite{curtarolo2012aflow} exhibit a consistent pipeline:
\textit{parse → normalize → encode → shard}

Processed materials datasets, such as those from AFLOW and OMat24, are often used to train graph neural networks like HydraGNN~\cite{pasini2024scalable}, which require encoding atomic structures as graph representations.
OMat24 offers over 100 million labeled DFT-derived structures, stored in a format compatible with large-scale ML pipelines. AFLOW automates the processing of vibrational, thermodynamic, and electronic descriptors and exposes these through structured APIs for ML-ready retrieval.
Unlike climate or fusion workflows, these representations are uniquely suited for graph-based learning. 

These datasets are often large, well-annotated, and well-suited to supervised learning—yet challenges remain in integrating multi-fidelity sources (e.g., experimental vs. simulation), managing class imbalance, and ensuring standardization across formats.



\begin{table*}[h]
\centering
\caption{2D conceptual maturity matrix integrating Data Readiness Levels and Data Processing Stages (grey cells N/A)}
\label{tab:readiness_matrix}
\adjustbox{max width=\columnwidth}{
\begin{tabular}{|l|p{2.5cm}|p{2.5cm}|p{2.5cm}|p{2.5cm}|p{2.5cm}|}
\hline
\textbf{Level} & \textbf{Ingest} & \textbf{Preprocess} & \textbf{Transform} & \textbf{Structure} & \textbf{Shard} \\
\hline
\textbf{1 - Raw} &
Initial raw acquisition &
\cellcolor{gray!10} &
\cellcolor{gray!10} &
\cellcolor{gray!10} &
\cellcolor{gray!10} \\
\hline
\textbf{2 - Cleaned} &
Validated ingestion into standard formats &
Initial spatial/temporal alignment or regridding &
\cellcolor{gray!10} &
\cellcolor{gray!10} &
\cellcolor{gray!10} \\
\hline
\textbf{3 - Labeled} &
Enhanced metadata enrichment &
Refined alignment; grids standardized &
Initial normalization or anonymization; basic labels added &
\cellcolor{gray!10} &
\cellcolor{gray!10} \\
\hline
\textbf{4 - Feature-engineered} &
Optimized high-throughput ingestion &
Alignment fully standardized &
Normalization or anonymization finalized; comprehensive labeling &
Domain-specific feature extraction completed &
\cellcolor{gray!10} \\
\hline
\textbf{5 - Fully AI-ready} &
Ingestion pipelines fully automated and performance-optimized &
Alignment integrated and automated &
Normalization / anonymization fully automated and audited &
Feature extraction automated and validated &
Data partitioned into train/test/val \& sharded into binary formats for scalable ingestion \\
\hline
\end{tabular}
}
\end{table*}

\subsection{Abstracted Workflow Patterns}

Across these four domains, we observe recurring patterns in the transformation of scientific data into AI-ready formats. While the specific workflows vary—from climate regridding to molecular graph encoding—common themes include spatiotemporal alignment, domain-informed feature extraction, data normalization, and sharding for high-performance I/O. These archetypes highlight that AI-readiness is not a binary state, but a spectrum shaped by domain constraints, data heterogeneity, and platform requirements. 


Each domain follows a characteristic pipeline composed of shared \textit{Data Processing Stages} and domain-specific transformations.
While readiness varies across domains and use cases, many workflows converge on a common set of technical transformations. We generalize these patterns into the following high-level pipeline:
\[
\textbf{ingest} \rightarrow \textbf{preprocess} \rightarrow \textbf{transform} \rightarrow \textbf{structure} \rightarrow \textbf{shard}
\]

\noindent This abstracted workflow introduces a finer distinction between \textit{transform} and \textit{structure}, which are often grouped in domain pipelines. The \textit{transform} stage captures domain-specific conversions (e.g., regridding in climate or anonymization in health), while \textit{structure} refers to organizing the data into standardized formats, such as fixed tensor layouts, hierarchical time-series, or graphs. The remaining stages—\textit{ingest}, \textit{preprocess}, and \textit{shard}—are broadly shared across domains and indicative of AI-readiness.

\section{Towards Scientific AI Readiness Framework}
\label{sec:readiness_framework}

Hiniduma et al.~\cite{hiniduma2025data} propose a 360-degree framework for data readiness that spans quality, accessibility, ethics, and infrastructure, offering a broad conceptual foundation. Building on this perspective, we introduce a complementary, compute-focused view rooted in operational workflows at DOE leadership facilities. In contrast to treating AI-readiness as a binary threshold, we formalize it as a multi-dimensional, technical process shaped by data origin, scientific objectives, and HPC constraints. 
To that end, we present five \textit{Data Readiness Levels}—raw, cleaned, labeled, feature-engineered, and fully AI-ready—that characterize how prepared datasets are for large-scale AI workflows. 
This system-level framework complements broader conceptual models by offering a pragmatic tool for evaluating technical readiness.
Still, we recognize that operational maturity alone does not capture the full complexity of scientific data preparation. To address this, Table~\ref{tab:readiness_matrix} integrates both the operational readiness levels and canonical scientific preprocessing stages into a unified conceptual data maturity matrix that holistically characterizes how datasets progress toward full AI-readiness.

Our analysis reveals that scientific AI-readiness is best understood through two complementary dimensions. The first is a set of Data Readiness Levels, which characterize the degree of technical preparation a dataset has undergone—from raw inputs to fully AI-trainable formats. The second is a set of domain-specific Data Processing Stages, which describe the structural workflow required to transform data within a particular scientific context. While the Data Readiness Levels provide a measure of readiness maturity, the patterns highlight the unique transformations required in different domains—such as regridding in climate science, anonymization in health data, or graph encoding in materials science. Together, these two axes offer a more complete framework for evaluating, comparing, and designing AI-ready data pipelines across the scientific enterprise. 
Looking ahead, advancing scientific AI-readiness at leadership scale may benefit from a set of guiding principles. These include: scalable preprocessing for large datasets, standardized formats and metadata for reproducibility, iterative pipelines with feedback loops, attention to governance and privacy, and alignment with HPC infrastructure for parallel training.

\section{Cross-Cutting Challenges}

Despite progress in developing AI-ready workflows across domains, several persistent challenges limit the reproducibility, scalability, and generalizability of scientific foundation models.

\textit{Data Scarcity and Quality Variability.}
Although DOE facilities often manage petascale datasets, many of them remain under-sampled, imbalanced, or inconsistent for ML training. Sparse labeling, non-standard metadata, and incomplete observational coverage reduce model robustness. Domain-specific maturity frameworks—such as METRIC~\cite{schwabe2024metric} for medical data or NOAA's climate data maturity model~\cite{bates2012maturitymodel}—provide useful guides but are rarely applied uniformly across scientific disciplines.

\textit{Scalability of Preprocessing Pipelines.}
Many AI-ready datasets require complex, multi-stage pipelines involving spatial/temporal regridding, physics-based feature extraction, and parallel I/O. These pipelines are often bespoke, domain-specific, and difficult to scale or share. Emerging workflow tools like SCWorks~\cite{kurihana2024scworks} show promise but need broader adoption and generalization to new modalities.

\textit{Provenance and Reproducibility.}
Establishing traceable links between raw data, preprocessing steps, and trained models is essential for validation, especially in open science settings. However, provenance capture remains ad hoc, with limited support for tracking transformations in HPC-scale workflows. Frameworks like ProvEn~\cite{souza2023provenance} at OLCF aim to address this, but broader integration into DRAI tooling is still needed.

\textit{Fragmentation Across Domains.}
Different scientific communities have developed AI-readiness pipelines in isolation, leading to inconsistent formats, terminology, and documentation. This fragmentation hinders transferability and collaboration. Leadership facilities are well-positioned to define common readiness templates, formats, and API-level standards that span disciplines.

\textit{Privacy, Security, and Compliance.}
Bio/health and national security datasets often contain PHI, PII, or CUI, requiring secure enclaves, auditability, and compliance with HIPAA or ITAR standards. These requirements complicate data preparation workflows and limit public model releases. NAIRR's secure enclave vision~\cite{OSTP_NAIRR_2023} represents an important step but requires practical implementation support at the facility level.

\textit{Data Quality, Bias, and Fairness.} 
Another important challenge, though not deeply explored here, involves ensuring data quality—addressing coverage, representativeness, imbalance, and noise. Approaches like Datasheets for Datasets or Data Cards can help identify potential biases and prevent issues like underfitting or overfitting. Incorporating feedback loops from model evaluation can further enhance data quality and model performance.

\section{Conclusions and Future Directions}

This paper highlights the diverse and evolving landscape of data readiness for AI across scientific domains. By analyzing real-world workflows in climate, fusion, bio/health, and materials science, we identify recurring preprocessing patterns and introduce a two-dimensional conceptual maturity matrix: one axis measuring Data Readiness Levels, and the other capturing domain-specific Data Processing Stages. This dual framework provides a practical foundation for benchmarking datasets, designing preprocessing infrastructure, and guiding cross-domain collaborations in scientific AI.

While some datasets achieve full AI-readiness—with sharded, normalized, and well-labeled inputs—others remain bottlenecked by domain-specific constraints such as sparse labels, inconsistent formats, and privacy limitations. These findings reinforce that readiness is not a binary state but a multi-dimensional process requiring both technical and contextual understanding.

Looking ahead, we envision the development of a reusable scientific AI-readiness framework composed of domain-specific templates, scalable preprocessing pipelines, provenance capture systems, and secure data enclaves. Such a framework would accelerate the training of scientific foundation models while supporting reproducibility, compliance, and operational efficiency across leadership-scale facilities. This readiness framework is especially relevant for transformer-based generative models, which increasingly shape the future of cross-domain scientific AI.

While this paper surveys key domains—climate, fusion, bio/health, and materials science—it does not capture the full breadth of scientific disciplines or the diversity of simulation codes in use today. Future work should broaden to more fields and tools, developing standardized domain-specific preprocessing templates for wider adoption.



\begin{acks}
This research used resources of the Oak Ridge Leadership Computing Facility at the Oak Ridge National Laboratory, which is supported under the Advanced Scientific Computing Research programs in the Office of Science of the U.S. Department of Energy under Contract No. DE-AC05-00OR22725. 
We would also like to thank Max Lupo Pasini, Jens Glaser, Yashika Ghai, Fernanda Foertter, John Gounley, and Heidi Hanson for helpful information regarding AI readiness challenges for specific domains. 
Finally, OpenAI's ChatGPT was used to provide editing suggestions for several sentences in the paper, as well as to help analyze some of the preprocessing patterns, given large chunks of code.
\end{acks}

\bibliographystyle{ACM-Reference-Format}
\bibliography{refs}

\end{document}